\documentclass[10pt,twocolumn,letterpaper]{article}

\usepackage{cvpr}
\usepackage{times}
\usepackage{epsfig}
\usepackage{graphicx}
\usepackage{amsmath}
\usepackage{amssymb}

\usepackage[pagebackref=true,breaklinks=true,letterpaper=true,colorlinks,bookmarks=false]{hyperref}

\cvprfinalcopy 

\def\cvprPaperID{2207} 

\newcommand{\SOA}[1]{\underline{\textbf{#1}}}

\ifcvprfinal\fi
\begin{document}

\title{Face Parsing with RoI Tanh-Warping}

\author{Jinpeng Lin$^{1}\thanks{Equal contribution. This work is done when Jinpeng Lin is an intern at Microsoft Research Asia.}$ ~~~ Hao Yang$^{2*}$ ~~~ Dong Chen$^{2}$ ~~~ Ming Zeng$^{1}\thanks{Corresponding author.}$ ~~~ Fang Wen$^{2}$ ~~~ Lu Yuan$^{2}$\\
 $^1$Software School of Xiamen University\\
 $^2$Microsoft Research\\
{\tt\small jplinforever@gmail.com  zengming@xmu.edu.cn} \\
{\tt\small  \{haya, doch, fangwen, luyuan\}@microsoft.com}
}

\maketitle

\begin{abstract}
Face parsing computes pixel-wise label maps for different semantic components (\eg, hair, mouth, eyes) from face images. 
Existing face parsing literature have illustrated significant advantages by focusing on individual regions of interest (RoIs) for faces and facial components. However, the traditional crop-and-resize focusing mechanism ignores all contextual area outside the RoIs, and thus is not suitable when the component area is unpredictable, \eg hair. 
%
%
Inspired by the physiological vision system of human, we propose a novel \emph{RoI Tanh-warping} operator that combines the central vision and the peripheral vision together.
It addresses the dilemma between a limited sized RoI for focusing and an unpredictable area of surrounding context for peripheral information.
%
%
To this end, we propose a novel hybrid convolutional neural network for face parsing. It uses hierarchical local based method for inner facial components and global methods for outer facial components.
The whole framework is simple and principled, and can be trained end-to-end. To facilitate future research of face parsing, we also manually relabel the training data of the HELEN dataset and will make it public. Experiments on both HELEN and LFW-PL benchmarks demonstrate that our method surpasses state-of-the-art methods.
\end{abstract}

\section{Introduction}

Given an input face image, face parsing assigns a pixel-wise label for each semantic component, \emph{e.g.}, hair, facial skins, eyes, nose, mouth and \emph{etc.}. Compared with face alignment, face parsing can provide more precise areas, and more importantly, face parsing can output the hair area, which is necessary for a variety of high level applications, such as face understanding, editing and animation.

Previous face parsing methods cannot handle hair segmentation very well. Some previous works~\cite{tsogkas2015deep,li2017integrated,gucclu2017end} crop a fixed-size area around the face and ignore the surrounding area. These methods often fail to obtain the whole hair area. \cite{liu2017face} propose using Fully Convolutional Networks (FCN) and Recurrent Propagation for face parsing, but it cannot distinguish the hair area of different people nearby. There are also some face parsing algorithms~\cite{wei2017learning,jackson2016cnn,luo2012hierarchical,zhou2015interlinked} that just ignore the hair area due to the limitations of the algorithm or the data.

The difficulty of the segmentation of both the inner facial components and the hair at one time is as follows. Generally, it is possible to improve accuracy by predicting each face region first, and then predict the per-pixel mask in each region individually. The estimated region yields good hints, including spatial constraint and instance information, for predicting high accuracy masks. However, if only focusing on the center of the face, the hair in the surrounding area will be ignored. 
This is a problem because we need an accurate and tight location of each face for inner facial components while requiring the surrounding context for the hair region.

\begin{figure}[t]
   \centering
   \includegraphics[width=\linewidth]{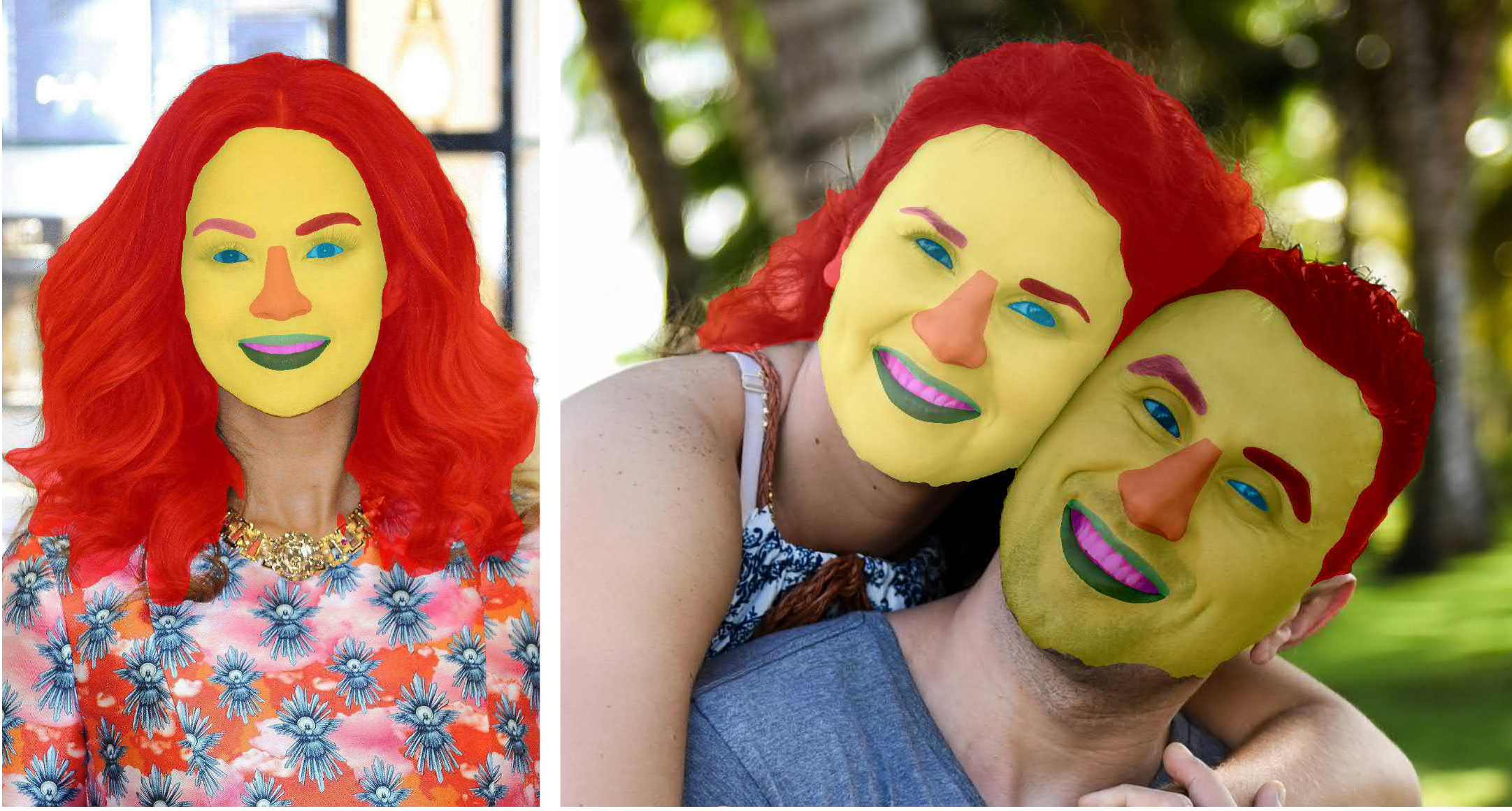}
   \caption{Our face parsing results. The proposed method is able to segment facial components as well as whole hair regions. It can also distinguish people who are close. Different brightness represent different individuals. (Best viewed in color)}
   \label{fig:intro}
    \vspace{-0.7cm}
\end{figure}

Our method is inspired by the human vision. When looking at something, human combine information of central vision and peripheral vision \cite{lettvin1976seeing}. While central vision covers narrow degrees of the visual field straight ahead of us, peripheral vision covers the rest. Peripheral vision is not as sharp as central vision, but has wider range and helps us detect events to the side, even when we're not looking in that direction. Based on this characteristic, we propose a novel RoI Tanh-warping operator which non-linearly maps the \emph{whole image} into a fixed-size. It addresses the dilemma between fixed input size and the unpredictable area of hair while reserving the amplified resolution on important regions.

Then, we feed the warped face image into a neural network. We use different strategies to process the inner and outer components of the face. For the inner components, \eg brows, eyes, nose and mouth, the network first predicts the bounding-box of each component, 
then maps the features of each component to a fixed size through RoI align~\cite{he2017mask}, a subnetwork is adopted to get the segmentation mask for each component. 
For the outer components, \eg hair, face region and background, we append a FCN~\cite{vaswani2017attention} to predict their masks.
Compared with existing local-based face parsing methods~\cite{luo2012hierarchical,zhou2015interlinked,liu2017face,gucclu2017end},
the proposed architecture reduces the computation cost greatly through sharing features and can be trained in an end-to-end way.

To the best of our knowledge, there are only a few public face parsing datasets, such as the HELEN~\cite{smith2013exemplar} and LFW part label (LFW-PL)~\cite{kae2013augmenting}. However, LFW-PL only labels 3 classes and the labeling of the HELEN's training data is not very precise, especially for hair. To facilitate future research of face parsing, we manually relabel the training data of HELEN. New labels are more accurate. We will publicize the new labellings, and hope it will attract more researcher to the topic of face parsing.

Without bells and whistles, our network surpasses previous state-of-the-art results on HELEN~\cite{smith2013exemplar} (trained on both new and old labellings) and LFW part label (LFW-PL)~\cite{kae2013augmenting} datasets. We summarize our contributions as follows:

\begin{enumerate}
\item We propose a novel method to address the problem of face parsing, for inner facial parts and hair. To the best of our knowledge, it is the first attempt to jointly solve the strongly related and severely imbalanced parts together, efficiently leveraging the spatial relationships between different parts.
\item We introduce a new \emph{RoI Tanh-warping} operation, which emphasizes the central face while retaining the peripheral parts (\ie, surrounding context) of the face, addressing the dilemma between fixed input size and unpredictable area of hair.
\item We devise a neural network which integrates a Mask R-CNN-fashion~\cite{he2017mask} branch and a FCN-fashion~\cite{long2015fully} branch to treat inner and outer facial parts, respectively. The hybrid method is simple, principled and can be trained end-to-end.
\item The results of our method surpasses all previous methods significantly on HELEN and LFW-PL dataset. We also relabel the original HELEN dataset~(mislabeled much) and plan to publicize it to facilitate future research. 
\end{enumerate}

\begin{figure*}[t]
\centering
\includegraphics[width=0.8\linewidth]{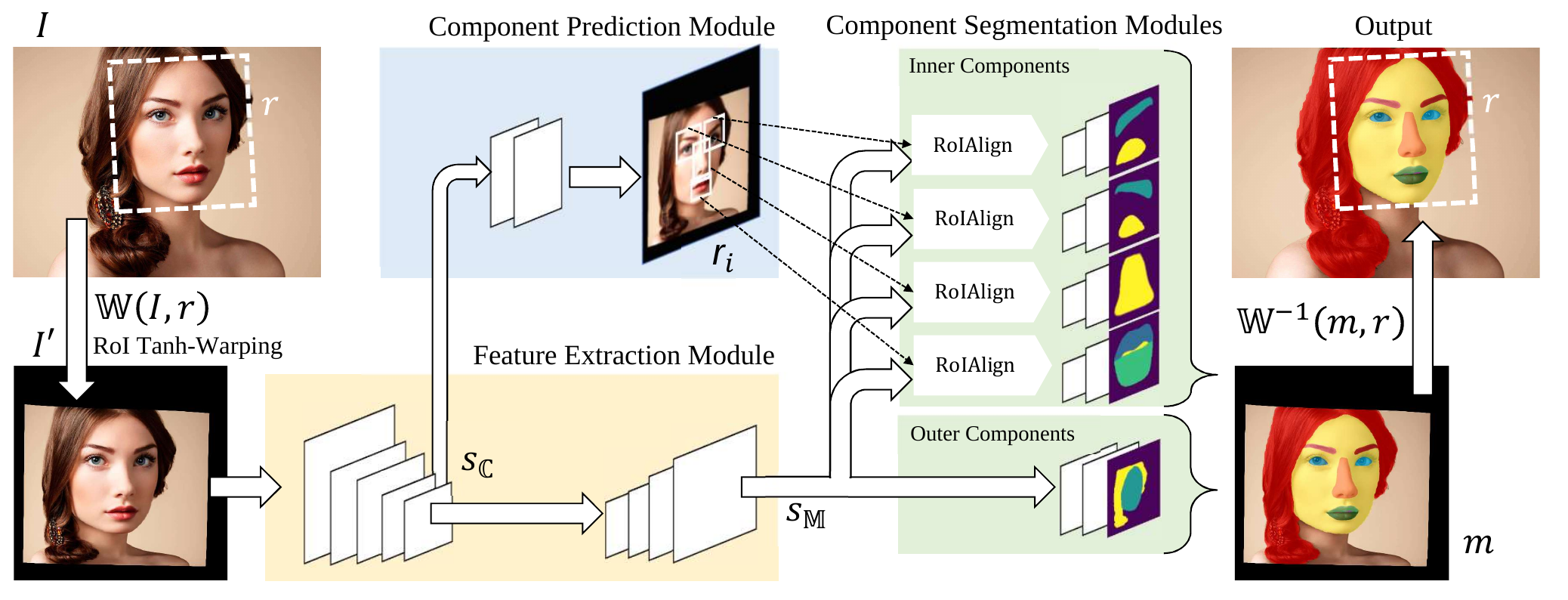}
\caption{Proposed network structure. Given an input image $I$, we retreive a face-aligned rectangle $r$, and apply RoI Tanh-warping $\mathbb{W}$ to retrieve a distorted and face-aligned image $I'$. We propose the \emph{feature extraction} module to extract feature maps from $I'$. For inner facial components (\eg, eye, nose, mouth and \etc), we use a \emph{component prediction} module to regress their local bounding boxes and use RoI align to extract their local features. For outer facial components (\ie, face, hair, background), we directly use the global feature. Then, for each component, we use an individual \emph{component segmentation module} to predict its segmentation scores though convolutions. The results are gathered as $m$ and warped back to the input domain as $\mathbb{W}^{-1}(m, r)$.}
\label{fig:network}
\vspace{-0.7em}
\end{figure*}

\section{Related Work}
\label{sec:relwork}

\textbf{Semantic Segmentation} Semantic segmentation for generic images has become a fundamental topic in computer vision, and achieved significant progress, e.g.\cite{BharathECCV2014,long2015fully,NohICCV2015,chen2016semantic,chen2016deeplab,szhengCRFRNNiccv2015,gao2016graph,niu2017feaboost,kwak2017weakly,he2017mask}. FCN~\cite{long2015fully} is a well-known baseline for generic images which employs full convolution on the entire image to extract per-pixel feature. Following this work, CRFasRNN~\cite{szhengCRFRNNiccv2015} and DeepLab~\cite{chen2016deeplab} adopt dense CRF optimization to refine the predicted label map. Hayder et al. \cite{hayder2017boundary} represent the segmentation mask as a truncated distance transform to alleviate the information loss caused by erroneous box cropping. Recently, Mask R-CNN~\cite{he2017mask} further advances the cutting edge of semantic segmentation through extending Faster R-CNN~\cite{renNIPS15fasterrcnn} and integrating a novel RoIAlign. However, directly applying these generic methods for face parsing may fail to model the complex-yet-varying spatial layout across face parts, especially hair, leading to unsatisfactory results. 

\textbf{Face Parsing} Most existing approaches for face parsing can be categorized into two groups: global-based and local-based methods.

Global-based methods directly predict per-pixel semantic label over the whole face image.  Early works represent spatial correlation between facial parts by various designed models, such as the epitome model~\cite{warrell2009labelfaces} and the exemplar-based method~\cite{smith2013exemplar}. With the advance of deep learning techniques, a variety of CNN structures and loss functions are proposed to encode the underlying layouts of the whole face. Liu et al. ~\cite{liu2015multi} integrate the CNN into the CRF framework, and jointly model pixel-wise likelihoods and label dependencies through a multi-objective learning method. Jackson et al.~\cite{jackson2016cnn} use facial landmarks as the guidance, and integrate boundary cue into CNN to implicitly confine facial regions. Zhou et al. ~\cite{zhou2017face} design an architecture which employs fully-convolutional network, super-pixel information, and CRF model jointly. Wei et al. ~\cite{wei2017learning} propose automatically regulating receptive fields in a deep image parsing network, therein obtaining better receptive fields for facial parsing. Besides these works, Saito et al. ~\cite{saito2016real} try to reduce computation to achieve real-time performance.

These kinds of methods treat face parts globally and inherently integrate the prior of the face layout. Nevertheless, accuracy is limited due to the lack of focusing on each individual part.

Local-based methods train separated models for various facial components (\eg eyes, nose \etc) to predict masks for each part individually. Luo et al. ~\cite{luo2012hierarchical} propose a hierarchical method which segments each detected facial part separately. Zhou et al. ~\cite{zhou2015interlinked} design an interlinked CNN-based pipeline which predicts pixel labels after facial localization. Benefiting from the complicated design, the interlinked CNN structure is able to pass information between coarse and fine levels bidirectionally, thus achieving good performance at the expense of large memory and computation consumption. Liu et al. ~\cite{liu2017face} achieve state-of-the-art accuracy with very fast running speed by combining a shallow CNN and a spatially variant RNN in two successive stages.

These local-based methods almost adopt the coarse-to-fine strategy, taking into account both global consistency and local precision. However, different from our proposed method, the previous methods model the two stages separately, without pursuing the gains of accuracy and efficiency from backbone sharing and joint optimization.

\textbf{Portrait/Hair Segmentation} Portrait segmentation and hair segmentation, such as the works of Shen et al.~\cite{Shen16EG, shenMatting16ECCV} and Chai et al.~\cite{Chai2016sigHair}, to name a few, are closely related to the literature of face parsing. Recent approaches for these two tasks adopt knowledge of specific domains into DCNN and achieve practical results for following up applications. Nevertheless, they only tackle a sub-problem of face parsing, without addressing the task of segmenting all parts on the face, while the latter is more general and challenging.

\section{Method}

We introduce the network with \emph{RoI Tanh-warping} for face parsing. Given a face image $I$ of varied dimensions, we first utilize a face and landmarks detector~\cite{chen2014joint} to retrieve five landmark points and construct a face rectangle $r$ based on the points. Then, we warp the whole image into fixed-sized $\mathbb{W}(I, r)$ with RoI Tanh-warping, where the face is aligned and centered within the output. The proposed RoI Tanh-warping operator $\mathbb{W}$ focuses on the aligned target face without loosing peripheral information. 

We use a hybrid solution to separately estimate masks for inner and outer facial components. For each inner facial component $\{P_i\}_{i=1}^N$ (\eg, eye, nose, mouth and \etc), where $N$ is the number of individual component, we predict the local rectangle $\{r_i\}_{i=1}^N$ bounding each component $P_i$ in the warped image. Then we extract local features within each rectangle $r_i$ through RoI align operators~\cite{he2017mask}. The pixel-wise segmentation scores $\{m_i\}_{i=1}^N$ for each component $P_i$ are predicted individually. For the outer facial components (\ie, hair, facial skin and background), we directly use the global feature $s_{\mathbb{M}}$ to predict the pixel-wise segmentation scores $m_{outer}$. Finally, we gather all scores denoted as $m$, and warp them back to the original input domain as $\mathbb{W}^{-1}(m, r)$ to get the final labels.

The whole pipeline of our method is shown in Figure~\ref{fig:network}. Our framework consists of four major components: 
1) the \emph{RoI Tanh-warping} operator $\mathbb{W}$, which warps input face image with varied dimensions into the face aligned and uniformly shaped domain;
2) the backbone \emph{feature extraction} module $\mathbb{F}$, which extracts informative features from the warped image for subsequent operations; 
3) the \emph{component prediction} module $\mathbb{R}$, which predicts the bounding rectangles of inner facial components; 
4) the multiple \emph{component segmentation} modules $\{\mathbb{M}_i\}$, which extract features from the predicted rectangles using RoI align operators for inner facial components or directly through up-samplings for outer facial components, and predict the pixel-wise segmentation scores. 

Unlike~\cite{luo2012hierarchical,zhou2015interlinked}, which separate the component prediction and segmentation into two individual steps, we share the common feature computation for component prediction and segmentation, which are jointly optimized and yield efficient computation. Next, we introduce each part in detail.

\subsection{RoI Tanh-Warping}
\label{sec:tanh-warping}

Previous face parsing works usually ignore the hair area, \eg \cite{wei2017learning,luo2012hierarchical,jackson2016cnn,zhou2015interlinked}, or only consider hair within a certain area, \eg \cite{smith2013exemplar,liu2015multi,li2017generative,li2017integrated}. This is because the area of the hair varies greatly, and the input to the network is usually a fixed-size picture of the face in the center. Fully Convolutional Networks (FCN)~\cite{long2015fully} can handle various input size. Unfortunately, the original FCN performing on the whole image treats the face globally. Its accuracy is limited due to lack of focusing on each individual part, for example, producing fragmentary masks for face parts.

\begin{figure}
\centering
\includegraphics[width=\linewidth]{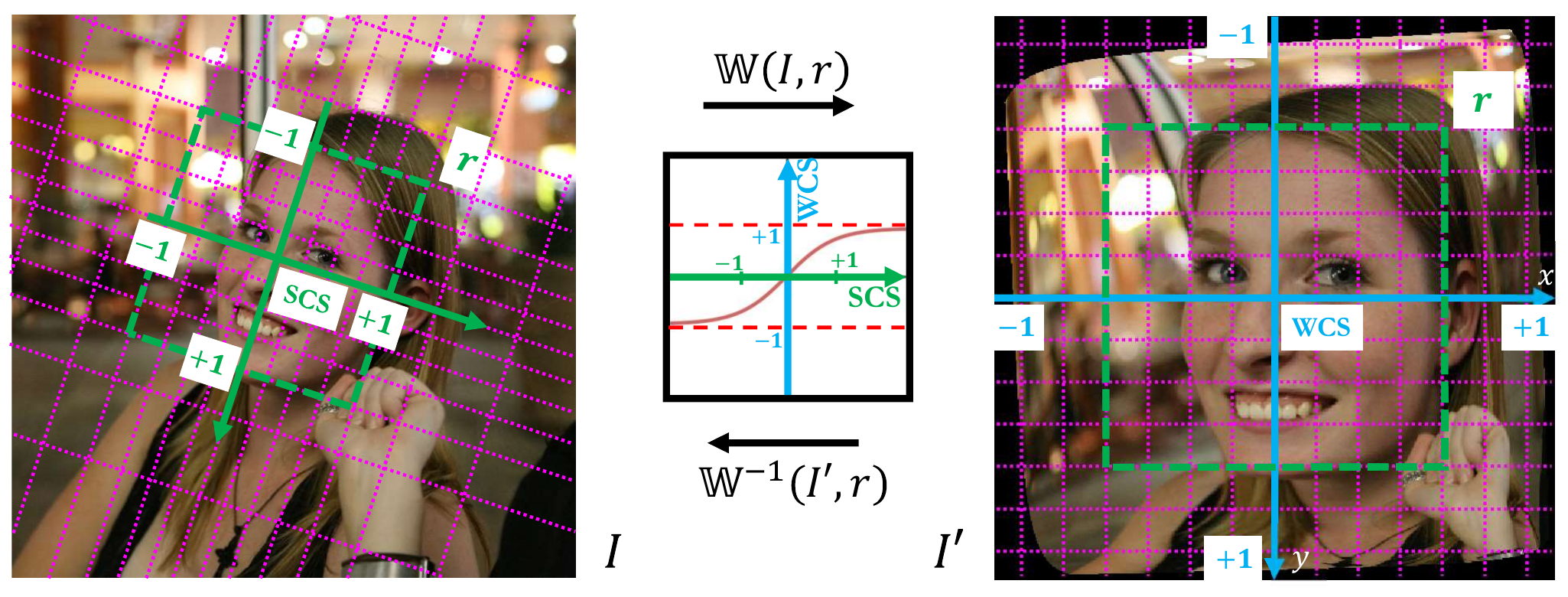}
\caption{
Essentially, Tanh-warping performs a $\tanh$ shaped distortion on x and y axes between two coordinate systems: the \emph{source coordinate system} (SCS, in green color) and the \emph{warped coordinate system} (WCS, in blue color) so that point located at $(x, y)$ in SCS in $I$ is warped to $(\tanh(x), \tanh(y))$ in WCS in $I'$. 
}
\label{fig:tanhwarp}
\vspace{-0.7em}
\end{figure}

To solve this problem, we propose RoI Tanh-warping, which maps the \emph{whole image of any size} into a limited view using a guiding rectangle. For convenience, we define two new coordinate systems, a \emph{source coordinate system} and a \emph{warped coordinate system}.

\noindent\textbf{Source Coordinate System}. Given an image $I$, we first apply a face and landmarks detector~\cite{chen2014joint} to get five landmarks representing the eyes, the nose and two corners of the mouth. We estimate a 2D similar transformation $T$ that transforms five detected landmarks to match five predefined template points\footnote{Predefined as $\{(-0.25, -0.1),(0.25, -0.1),(0, 0.1),(-0.15, 0.4),$ $(0.15,0.4)\}$.}. The face rectangle $r$ is then determined by the four corner points computed by $T^{-1}(\pm 1,\pm 1)$. With the rectangle $r$ in $I$, a local coordinate system can  be determined: its origin is the center of $r$, its $x=\pm 1$ and $y=\pm 1$ lines coincide with $r$'s borders. The source coordinate system is shown in the left image of Figure \ref{fig:tanhwarp}. 

\noindent\textbf{Warped Coordinate System}. Let $I'$ be the warped image, the warped coordinate system is defined as: its origin is the center of $I'$, while its $x=\pm 1$ and $y=\pm 1$ lines determine the borders of $I'$, as shown in right image of Figure \ref{fig:tanhwarp}.

Suppose a pixel $p$ in $I$ is warped to $p'$ in $I'$ using RoI Tanh-warping, then its original coordinates $(x,y)$ under the source coordinate system and the new coordinates $(x', y')$ under the warped coordinate system satisfy:
\vspace{-0.6em}
\begin{equation}
x'=\tanh(x), \text{~~~} y'=\tanh(y).
\vspace{-0.6em}
\label{equ:warping}
\end{equation}

\begin{figure}[t]
\centering
\includegraphics[width=\linewidth]{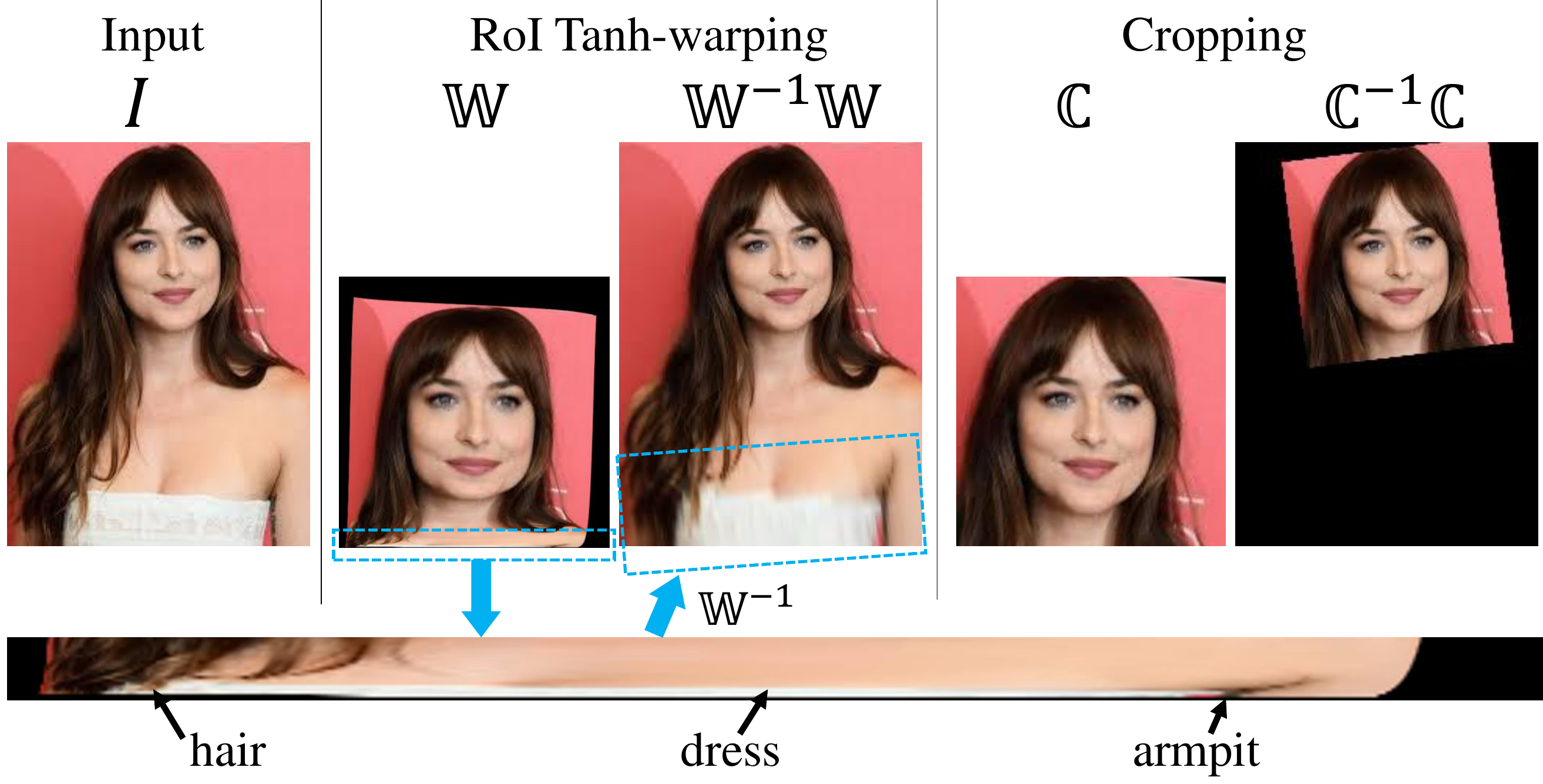}
\caption{Tanh-warping vs. cropping. The Tanh-warping $\mathbb{W}$ is better invertible than cropping $\mathbb{C}$ on peripheral area.
}
\label{fig:sigmoid_warp_invert}
\vspace{-0.7em}
\end{figure}

We denote the RoI Tanh-warping as $\mathbb{W}(I, r)$. It is implemented as a bilinear sampling according to Equation \ref{equ:warping}. In theory, $\mathbb{W}$ is invertible.
Let $\mathbb{W}^{-1}$ be the inverse operator of $\mathbb{W}$. We show results of $\mathbb{W}$ and $\mathbb{W}^{-1} \mathbb{W}$ on a face image $I$ in Figure \ref{fig:sigmoid_warp_invert}.
The proposed $\mathbb{W}(I,r)$ contains almost all information of the original image $I$, including the hair, dress and armpit in the bottom. 
It is observed that: i) although some fine details may be lost during the warping, the whole shape of hair can be basically recovered through $\mathbb{W}^{-1}$;
ii) the RoI Tanh-warping operator preserves more linearity near the center of the rectangle $r$, but imposes more distortion on the peripheral pixels (\ie surrounding context).

Most previous face parsing methods~\cite{liu2017face,zhou2017face,gucclu2017end} apply face alignment via cropping. It is hence infeasible for these approaches to predict the labels (\eg, hair) on peripheral pixels distant from the face, as shown in Figure \ref{fig:sigmoid_warp_invert}. The proposed RoI Tanh-warping addresses the dilemma between fixed input size and unpredictable area of hair while reserving the focusing on the RoI.

\subsection{Feature Extraction}

Given the warped face image $\mathbb{W}(I,r)$, which contains only a single face in the center of the image, the \emph{feature extraction} module $\mathbb{F}$ is deployed to capture implicit features shared by multiple tasks. $\mathbb{F}$ uses the ResNet-18 and FPN~\cite{lin2017feature} structure as the backbones.
FPN helps achieve a higher spatial resolution through layer-by-layer feature upsampling since the output feature map from C4 (output of the conv4 block of the ResNet-18) has very low resolution, which is too coarse for accurate segmentation. We can get the feature maps $s_{\mathbb{R}}$ and $s_{\mathbb{M}}$ for component prediction and segmentation respectively by
\vspace{-0.6em}
\begin{equation}
(s_{\mathbb{R}}, s_{\mathbb{M}}) = \mathbb{F}(x, \theta_\mathbb{F}),
\vspace{-0.6em}
\end{equation}
where $\theta_\mathbb{F}$ denotes the parameters of $\mathbb{F}$, including the parameters of the ResNet and the FPN. $s_{\mathbb{R}}$ represents the C4 layer of ResNet-18. $s_{\mathbb{M}}$ represents the last layer of FPN.

\subsection{Component Prediction}
\label{ssec:component_prediction}

\begin{figure}
\centering
\includegraphics[width=0.49\linewidth]{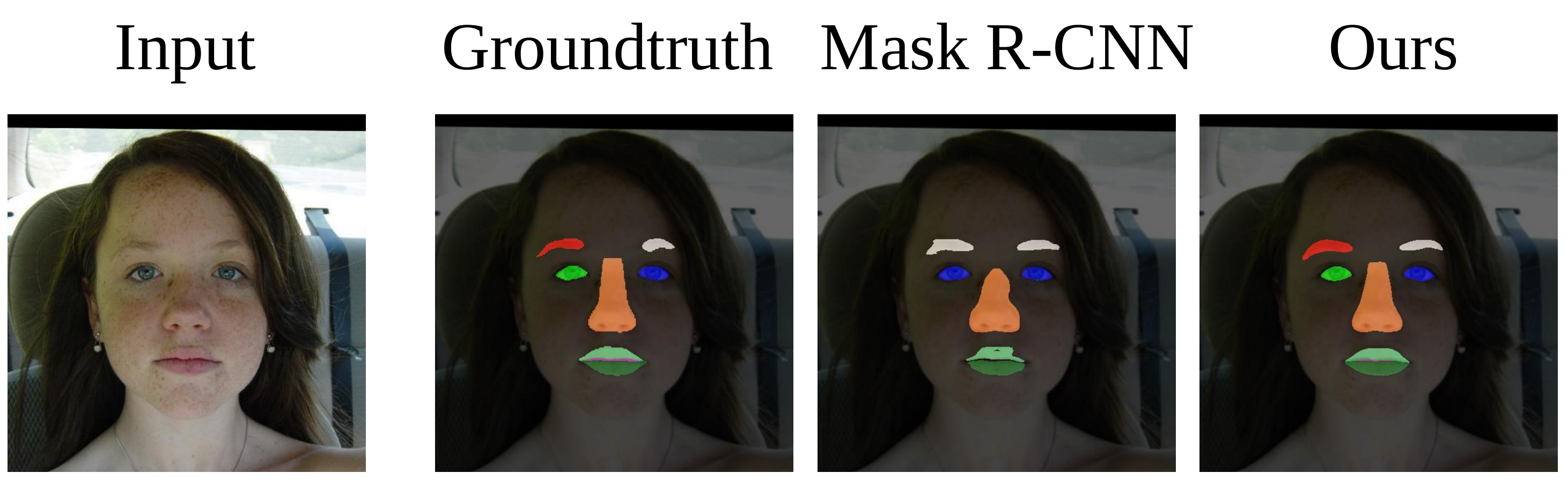}
\includegraphics[width=0.49\linewidth]{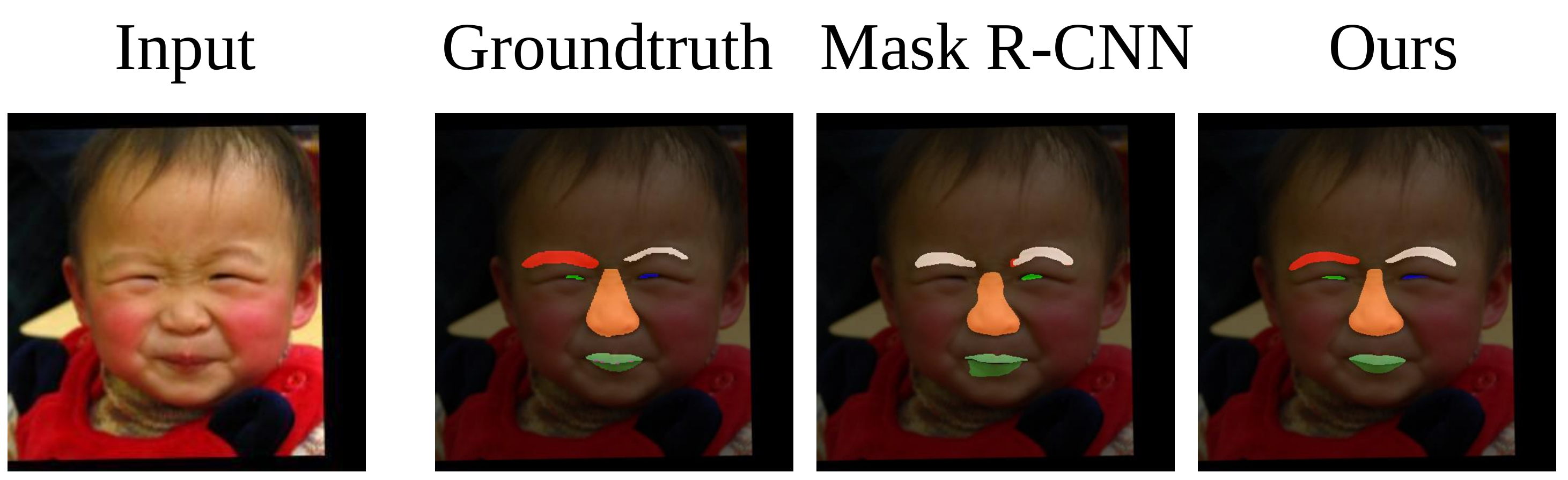}
\caption{Directly applying Mask R-CNN in face parsing causes problems: the left brows and left eyes are recognized as the right brows/eyes. The region proposal network that Mask R-CNN relies on misclassified instances that share similar appearance but have different semantic labels. The box regression we use is more straight-forward but effective for parsing facial components.}
\label{fig:mrcnn}
\vspace{-0.6em}
\end{figure}

The \emph{component prediction} module $\mathbb{R}$ locates bounding rectangles of the $N$ inner facial components: $\{r_i\}_{i=1}^N$. It directly outputs the coordinates of each rectangle, namely
\vspace{-0.6em}
\begin{equation}
\{r_i\}_{i=1}^N = \mathbb{R}(s_{\mathbb{R}}, \theta_\mathbb{R}),
\vspace{-0.6em}
\end{equation}
where $N$ is the number of inner facial components, and $\theta_\mathbb{R}$ is the parameters of $\mathbb{R}$.
The \emph{component prediction} module $\mathbb{R}$ consists of two convolutional layers followed by a global average pooling and a fully connected layer. We consider a tight rectangle surrounding the annotated component mask as the ground-truth bounding-box, denoted as $\{r_i^g\}_{i=1}^N$. We adopt the $L_1$ loss for the bounding-box regression: 
\vspace{-0.6em}
\begin{equation}
\mathcal{L}_{comp}= \frac{1}{N}\sum_{i=1}^N \|r_i - r_i^g\|_1.
\label{equ:comploss}
\vspace{-0.6em}
\end{equation}
Here we explicitly regress the area of each component instead of detecting them individually like in Mask R-CNN. The semantic label of every predicted facial component is explicitly defined. It avoids ambiguities in components and reduces computation cost, 
as shown in Figure \ref{fig:mrcnn}.


\subsection{Component Segmentation}

We proposed a hybrid structure for component segmentation. For each inner facial component (eyes, brows, nose, mouth), we use RoI align to extract the local feature and predict its masks individually. The estimated region in previous step yields good hints for predicting high accuracy masks. For the outer facial components (face, hair), we use FCN to directly predict the segmentation mask in the warped face. It can handle unpredictable area of hair.

\noindent\textbf{Inner Facial components}. For the obtained $N$ bounding boxes, $N$ light and parallel component segmentation modules $\{\mathbb{M}_i\}_{i=1}^N$ are applied to predict the pixel-wise masks for each inner facial component. All component segmentation modules share the same network architecture, but their weights are independent. Without loss of generality, we only discuss the $i^\textit{th}$ segmentation network as follows.


Given a bounding-box $r_i$ regressed from $\mathbb{R}$, we apply RoI-align operator to sample a local feature patch out from the feature map $s_{\mathbb{M}}$ and resize it to a squared size. We observe that sometimes the regressed bounding-box is not very accurate: some areas may fall outside the box, especially for the wide open mouth. This may be caused by insufficient training data, but if it happens, the area outside the box will not be correctly segmented. So we add a padding outside the box to solve this problem. The RoI align feature $l_i$ for the $i^{\textit{th}}$ component is defined as
\vspace{-0.6em}
\begin{equation}
l_i = \texttt{RoIAlign}(s_\mathbb{M}, \texttt{Padding}(r_i)).
\label{equ:tanhpooling}
\vspace{-0.6em}
\end{equation}
Then we perform several convolutions and up-sampling operations to generate a segmentation $m_i$ of the $i^{\textit{th}}$ component.
\vspace{-0.6em}
\begin{equation}
m_i = \mathbb{M}_i(l_i, \theta_{\mathbb{M}_i})
\label{equ:mask_pred}
\vspace{-0.6em}
\end{equation}
where $\theta_{\mathbb{M}_i}$ represents the parameters of the $i^{\textit{th}}$ segmentation module. We use the pixel-wise cross-entropy to measure the component segmentation accuracy. The segmentation loss $\mathcal{L}_{inner}$ is defined as the averaged cross-entropy among all the segmentation networks:
\vspace{-0.6em}
\begin{equation}
\mathcal{L}_{inner} = \frac 1 {N} \sum_{i=1}^{N} \texttt{CrossEntropy}(m_i, m_i^g),
\label{equ:maskloss}
\vspace{-0.6em}
\end{equation}
where $m_i^g$ is the ground-truth segmentation of the $i^\textit{th}$ component. 

\noindent\textbf{Outer Facial Components}. For the outer facial components (\ie, hair, facial skin and background), we apply FCN to get the segmentation mask, as shown in Figure~\ref{fig:network}.
\vspace{-0.6em}
\begin{equation}
m_{outer} = \mathbb{M}_{outer}(s_\mathbb{M}, \theta_{\mathbb{M}})
\vspace{-0.6em}
\end{equation}
We also use the cross-entropy loss to constrain the segmentation accuracy:
\vspace{-0.6em}
\begin{equation}
\mathcal{L}_{outer} = \texttt{CrossEntropy}(m_{outer}, m_{outer}^g),
\label{equ:outmaskloss}
\vspace{-0.6em}
\end{equation}
where $m_{outer}^g$ is the ground-truth segmentation of the outer facial component. 

Finally, all the resulting segmentation scores are gathered, denoted as $m$. We de-warp the scores to the original image domain as $\mathbb{W}^{-1}(m, r)$ using the same rectangle $r$ from the input stage and form the final face parsing result.

\subsection{Implementation Details}
\label{ssec:implementation}

Since the component segmentation relies on a good component region estimation, we divide the training process into two stages. In the first stage, we only train the \emph{feature extraction} module and the \emph{component prediction} module for a good component regressing accuracy. Here, only the component loss $\mathcal{L}_{comp}$ (in Equation \ref{equ:comploss}) is used for training. In the second stage, we perform joint training by updating all parameters, including $\theta_{\mathbb{F}}$, $\theta_{\mathbb{R}}$, and $\theta_{\mathbb{M}}$, with the component loss $\mathcal{L}_{comp}$ (in Equation \ref{equ:comploss}) and the mask loss $\mathcal{L}_{inner}$ and $\mathcal{L}_{outer}$ (in Equation \ref{equ:maskloss} and \ref{equ:outmaskloss}) together.

By default, the size of warped images $I'$ is set to $512\times 512$. We use ResNet-18 in \emph{feature extraction}. We select the feature from $C_4$ of ResNet as $s_{\mathbb{R}}$ for component prediction, and the feature from $P_2$ of FPN as $s_{\mathbb{M}}$ for component segmentation. 
The spatial resolution of $s_{\mathbb{R}}$ is $32 \times 32$, the spatial resolution of $s_{\mathbb{M}}$ is $128 \times 128$. The \emph{component prediction} module consists of two $3 \times 3 \times 320$ convolutional layers, one $1 \times 1 \times 1280$ convolutional layer, one global average pooling and one fully connected layer. Its output is a tensor of $N \times 4$ that encodes the location of $N$ bounding boxes, where $N$ is the number of inner facial components. $N$ varies for different databases. Before RoI align, regressed boxes are padded by $10\%$ the feature map size for mouth and $5\%$ otherwise. The output size of RoI align defined by Equation \ref{equ:tanhpooling} is $32 \times 32$ for all inner components. Each \emph{component segmentation} module is built with two $3\times 3 \times 256$ convolutions each followed by one bilinear up-sampling if for inner component. A $1 \times 1$ convolution followed by softmax is appended to each \emph{component segmentation} module to output the masks. The sizes of masks $\{m_i\}$ are all $128 \times 128$. 

  
\section{Experiments}
\label{sec:experiments}


We use HELEN~\cite{smith2013exemplar} and LFW-PL~\cite{kae2013augmenting} for the experiments. The HELEN dataset contains 2,330 face images. Each image is annotated with 11 labels: {``background"}, {``facial skin"}, {``left/right brow"}, {``left/right eye"}, {``nose"}, {``upper/lower lip"}, {``inner mouth"} and {``hair"}. We adopt the same dataset division setting as in~\cite{liu2015multi,yamashita2015cost,wei2017learning} that uses 2,000 images for the training, 230 images for the validation and 100 images for the testing. The LFW-PL dataset contains 2,972 face images. All of them are manually annotated with 3 labels: {``skin"}, {``hair"} and {``background"}. Following~\cite{kae2013augmenting,zhou2017face}, we use 1,500 images for the training, 500 images for the validation, and 927 images for the testing.

\begin{figure}[b]
   \centering
   \includegraphics[width=\linewidth]{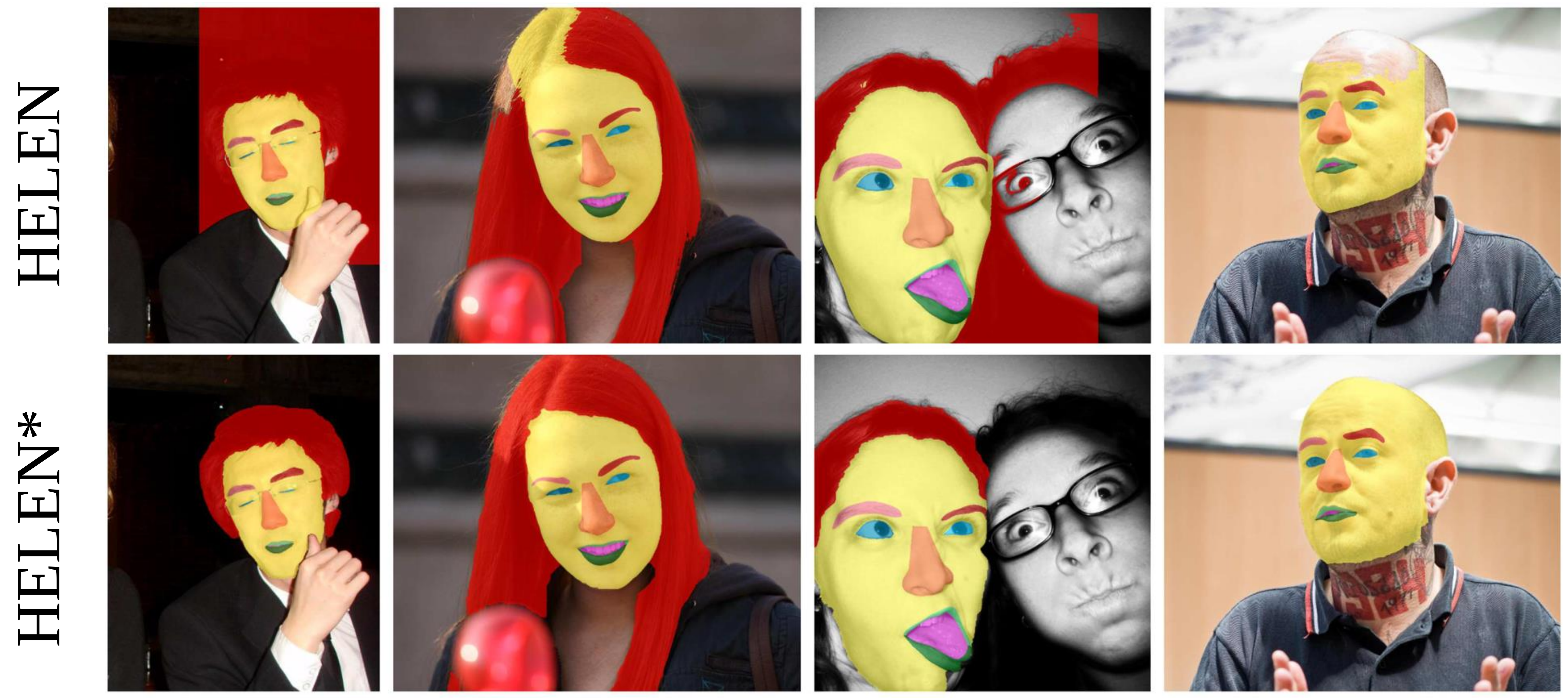}
   \caption{HELEN vs. HELEN*. Obvious annotation errors of face and hair in HELEN are all rectified in HELEN*.}
   \label{fig:hair_annotation}
  \vspace{-0.7em}
\end{figure}

Due to the inaccurate annotations in HELEN, we re-annotate its ``hair" and ``facial skin" labels manually. We also reprocess the facial component labels according to the modified facial skin label. For fair comparison with the previous methods, we only re-annotate both training and validation sets, and the test set remains unchanged. We refer to this relabeled dataset as HELEN*. Figure \ref{fig:hair_annotation} compares the annotations between HELEN and HELEN*.

We augment data during training: 1) randomly replace the background with non-face images or pure colors, 2) random rotation (within $[-18^\circ,18^\circ]$), scaling (within $[0.9, 1.1]$) around the face center, 3) random horizontal flipping and 4) random gamma adjustment with $\gamma \in [0.5, 2]$.

\subsection{Comparison with State-of-the-art}

We perform a thorough comparison between our model and existing state-of-the-art methods on HELEN and LFW-PL datasets. Results are measured by F-measure which is commonly used by existing face parsing literature. Our results are calculated using the original image sized annotations without any transformation or cropping.

We show the comparison results on HELEN in Table \ref{table:helen_sta}. Each column shows the F-measure percentage corresponding to a specific face label. \emph{I-mouth} is short for inner mouth, \emph{U/L-lip} is short for \emph{upper/lower lip}, and \emph{overall} represents a union of all inner facial component (eyes/brows/nose/mouth) labels.
We report the results of our model trained on both HELEN and HELEN* in Table \ref{table:helen_sta}. The testing results on HELEN show that our method outperforms all state of the art methods. We also observe an improvement brought by training on the relabeled HELEN* dataset, especially on the skin and hair labels.

\vspace{-2mm}
\begin{table}[h]
   \begin{center}
      \scriptsize
      \addtolength{\tabcolsep}{-5pt}
      \begin{tabular}{l|cccccccccc}
         Methods              & eyes       & brows      & nose       & I-mouth    & U-lip      & L-lip  & mouth      & skin       & hair & overall  \\
         \hline
         Smith \etal \cite{smith2013exemplar}   & 78.5       & 72.2       & 92.2       & 71.3       & 65.1       & 70.0   & 85.7       & 88.2       & -    & 80.4       \\
         Zhou \etal \cite{zhou2015interlinked} & 87.4       & 81.3       & 95.0 & 83.6       & 75.4       & 80.9   & 92.6       & -          & -    & 87.3       \\
         Liu \etal \cite{liu2015multi}     & 76.8       & 71.3       & 90.9       & 80.8       & 62.3       & 69.4   & 84.1       & 91.0       & -    & 84.7       \\
         Liu \etal \cite{liu2017face}      & 86.8       & 77.0       & 93.0       & 79.2       & 74.3       & 81.7   & 89.1       & 92.1       & -    & 88.6       \\
         Wei \etal \cite{wei2017learning}  & 84.7       & 78.6       & 93.7       & -          & -          & -      & 91.5       & 91.5       & -    & 90.2       \\
         \hline
         Ours ({\tiny trained on HELEN}) &  89.6 & 83.1  & 95.6 & 86.7    & 79.6  & \SOA{89.8}  & 95.0  & 94.5 & 83.5 & 92.4 \\ 
         Ours ({\tiny trained on HELEN*}) & \SOA{89.7} & \SOA{85.9}  & \SOA{95.6} & \SOA{86.7}  & \SOA{80.8}  & 89.7  & \SOA{95.2}  & \SOA{95.3} & \SOA{88.7} & \SOA{93.1} \\
      \end{tabular}
   
   \end{center}
   \vspace{-0.5em}
   \caption{Comparison with state-of-the-art methods on HELEN.}
    \label{table:helen_sta}
\end{table}

We also show the comparison between our model and existing face parsing methods on LFW-PL in Table \ref{table:lfw_sta}. We report the F-measure percentages corresponding to skin, hair and background. We compare the overall accuracies as well. Our method also surpasses state of the art on the LFW-PL dataset. The improvement of our method is relatively small, since all images in LFW-PL have been aligned and cropped, only one face is in the middle of the image.

\begin{table}[h]
   \begin{center}     
      \scriptsize
      \begin{tabular}{l|ccc|c}
         Methods                          & skin & hair & bg & accuracy  \\
         \hline
         Liu \etal \cite{liu2015multi}             & 93.93 & 80.70  & 97.10 & 95.12 \\
         Long \etal \cite{long2015fully}           & 92.91 & 82.69  & 96.32 & 94.13 \\
         Chen \etal \cite{chen2016deeplab}         & 92.54 & 80.14  & 95.65 & 93.44 \\
         Chen \etal \cite{chen2016semantic}        & 91.17 & 78.85  & 94.95 & 92.49 \\
         Zhou \etal \cite{zhou2017face}            & 94.10 & 85.16  & 96.46 & 95.28 \\
         Liu \etal \cite{liu2017face}              & \SOA{97.55} & 83.43  & 94.37 & 95.46 \\
         \hline
         Ours & 95.77 & \SOA{88.31} & \SOA{98.26} & \SOA{96.71}  \\
      \end{tabular}
   \end{center}
      \vspace{-0.5em}
    \caption{Comparison with state-of-the-art methods on LFW-PL. }
      \label{table:lfw_sta}
     \vspace{-1em}
\end{table}

\begin{table*}[t]
\begin{center}

\scriptsize
\addtolength{\tabcolsep}{-3pt}
\begin{tabular}{c|c|cccccccccc}
input & network structure      & eyes & brows & nose & I-mouth & U-lip & L-lip & mouth & skin & hair & overall  \\
\hline
rescale & FCN              & 77.5 & 66.0  & 69.7 & 71.4    & 62.7  & 68.4  & 79.9  & 80.3 & 82.7 & 73.0     \\ 

\hline
$\mathbb{C}$ & FCN           & 82.7 & 79.6  & 93.7 & 86.3    & 78.1  & 85.2  & 92.5  & 94.4 & 85.0 & 89.4     \\ 
$\mathbb{C}$ & Hybrid w/o Padding  & 88.6 & 83.9  & 94.5 & \SOA{87.6}    & 79.4  & 89.7  & 94.5  & 95.1 & 84.6 & 92.0  \\ 
$\mathbb{C}$ & Hybrid  & 89.7 & 84.7 & 95.5 & 86.9    & 80.3  & 90.7  & \SOA{95.4}  & 95.2 & 84.5 & 92.9 \\ 

\hline

$\mathbb{C}2$ & Hybrid & 85.9 & 83.3 & 94.4 & 84.1 & 73.6 & 85.5 & 92.5 & 91.7 & 83.8 & 90.9 \\  

\hline
$\mathbb{W}$ & FCN             & 82.6 & 79.3  & 93.8 & 85.7    & 77.0  & 84.4  & 92.4  & 94.0 & \SOA{88.8} & 89.3     \\ 
$\mathbb{W}$ & Hybrid w/o Padding     & 88.1 & 84.7  & 94.6 & 87.4    & 76.7  & 89.6  & 94.1  & 95.0 & 88.4 & 91.9     \\ 

$\mathbb{W}$ & Hybrid w/ WeightSharing & 89.6 & 85.1 & 95.6 & 85.4     & 75.8  & 89.2  & 94.3  & 94.9 & 88.1 & 92.7 \\ 

$\mathbb{W}$ & Hybrid w/ EyeBrowSymmetry & 89.7 & 85.8 & 95.5 & 86.8 & 80.5 & 89.6 & 95.1 & 95.3 & 88.5 & 93.0 \\ 

$\mathbb{W}$ & Hybrid (Ours)     & \SOA{89.7} & \SOA{85.9}  & \SOA{95.6} & 86.7    & \SOA{80.8}  & \SOA{89.7}  & 95.2  & \SOA{95.3} & 88.7 & \SOA{93.1}     \\ 

\end{tabular}
\end{center}
\vspace{-0.5em}
\caption{Comparison of ablation models trained on HELEN*.} 
\label{table:compare_with_baseline_helen_star}
\vspace{-0.5em}
\end{table*}

\subsection{Ablation Study}
\label{ssec:ablation}

\begin{figure}
\centering
\includegraphics[width=\linewidth]{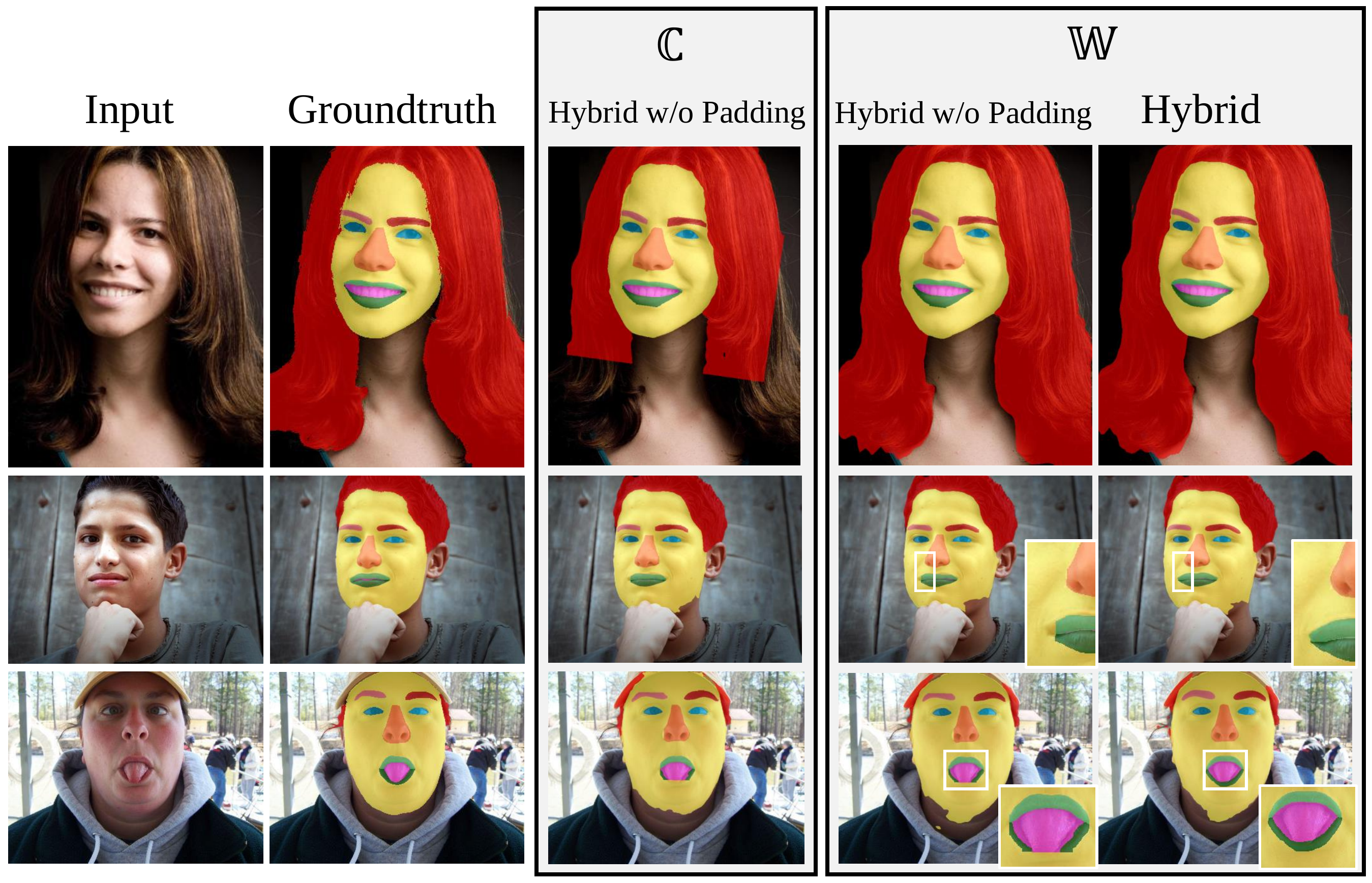}
\caption{Qualitative comparisons on HELEN dataset. The proposed Tanh-warping $\mathbb{W}$ addresses the hair cropping issues; the padding in our hybrid structure is also necessary when box regression is not accurate.}
\label{fig:ablation_results}
\vspace{-0.7em}
\end{figure}

To understand the role of the hybrid network structure and the proposed RoI Tanh-warping, we conduct several baseline methods for comparisons. 
We substitute certain submodules of our proposed framework with alternatives and construct four network structures, which are:

\noindent\textbf{FCN:} An FCN structure without the component prediction module $\mathbb{R}$. All segmentation modules $\mathbb{M}_i$ directly take $s_{\mathbb{M}}$ as the input feature without using RoI-align.

\noindent\textbf{Hybrid w/o Padding:} A hybrid structure with component prediction module applied to regress bounding boxes of inner facial components. Segmentation modules for inner components all take the RoI-aligned features as $\texttt{RoIAlign}(s_\mathbb{M}, r_i)$, but without any padding on the bounding boxes $r_i$.

\noindent\textbf{Hybrid:} The proposed hybrid structure. Segmentation modules for inner components take RoI-aligned features as inputs. The bounding boxes for RoI-align are padded  to better tolerate box regression errors following Equation \ref{equ:tanhpooling}.

\noindent\textbf{Hybrid w/ WeightSharing:} A hybrid structure similar to our proposed model, except that weight sharing is applied to all its segmentation modules for inner components (except for the last $1\times 1$ convolution before soft-max). We use this baseline model to simulate the mask heads of Mask R-CNN. 

\noindent\textbf{Hybrid w/ EyeBrowSymmetry:} The parameters $\theta_{\mathbb{M}_i}$ are shared between branches of left/right eyes and branches of left/right brows. Specifically, features $l_i$ cropped from Equation \ref{equ:tanhpooling} for the \emph{left eye} and \emph{left brow} labels are horizontally flipped before forwarding to the component segmentation modules $\mathbb{M}_i$. In return, the output segmentations $m_i$ from Equation \ref{equ:mask_pred} for the \emph{left eye} and \emph{left brow} labels are flipped back to compose the final prediction. 

We combine these network structures with different techniques that help focus the network on the input: 1) cropping the image within the face rectangle $r$, denoted by $\mathbb{C}$; 2) Tanh-warping the image guided by $r$, denoted by $\mathbb{W}$; 3) rescaling all the input images to $512\times 512$ (with padding to preserve the aspect ratios), denoted by $\texttt{rescale}$.

Eight baseline models are listed by Table \ref{table:compare_with_baseline_helen_star}, including the proposed model $\mathbb{W}$+\emph{Hybrid}. All the models are trained on HELEN*.
We compare the proposed model with other baseline methods. Some visual results are compared in Figure \ref{fig:ablation_results}. 
It shows the improvement brought by the Tanh-warping operator in parsing hairs, and the advantage of our hybrid network structure in parsing inner facial components.
From these results, we can draw some conclusions:

\begin{figure}[b]
\centering
\includegraphics[width=\linewidth]{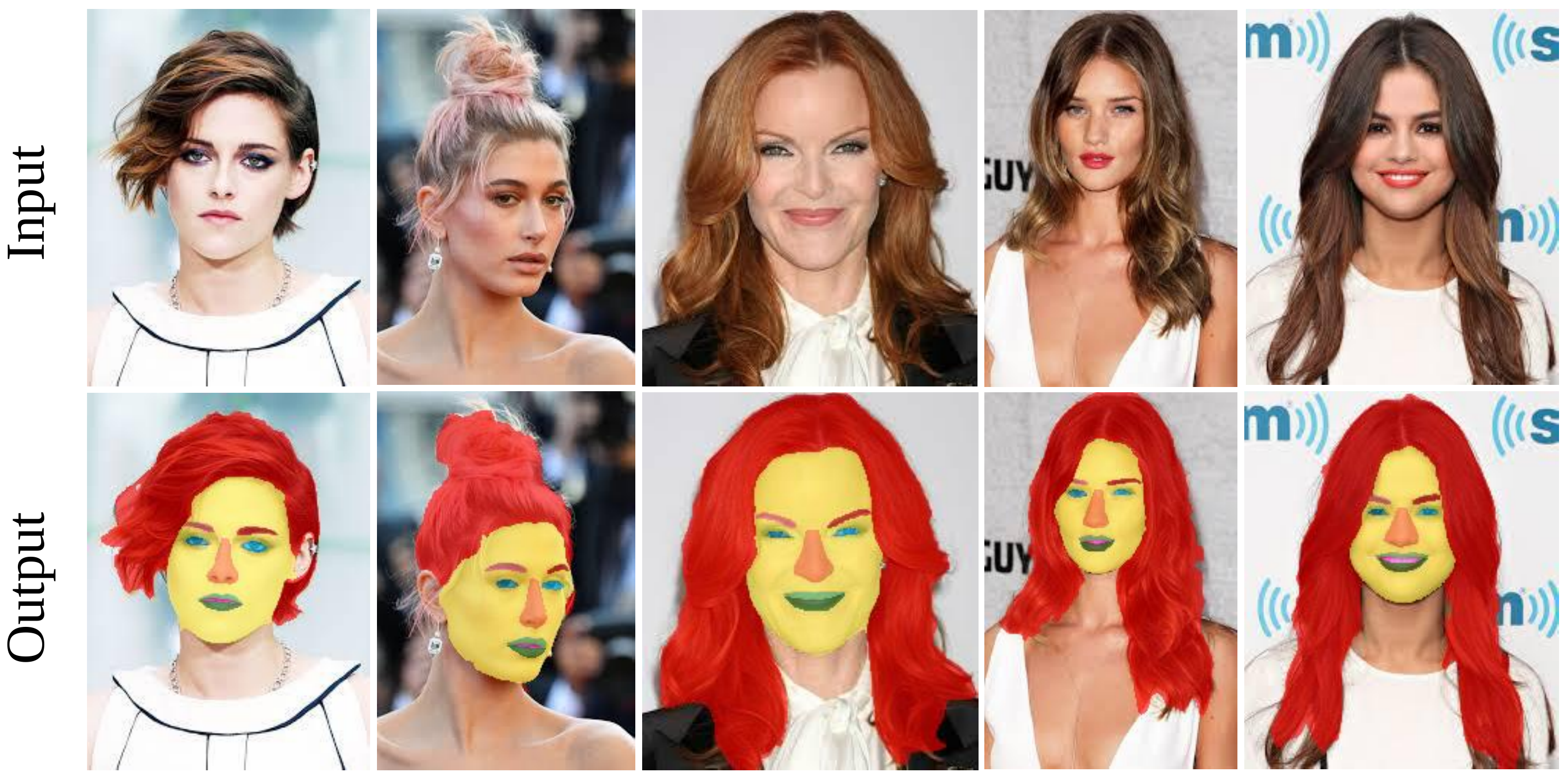}
\caption{Our method can handle hairs with various lengths.}
\label{fig:hairs}
\vspace{-0.7em}
\end{figure}

\begin{figure*}[t]
\centering
\includegraphics[width=0.99\linewidth]{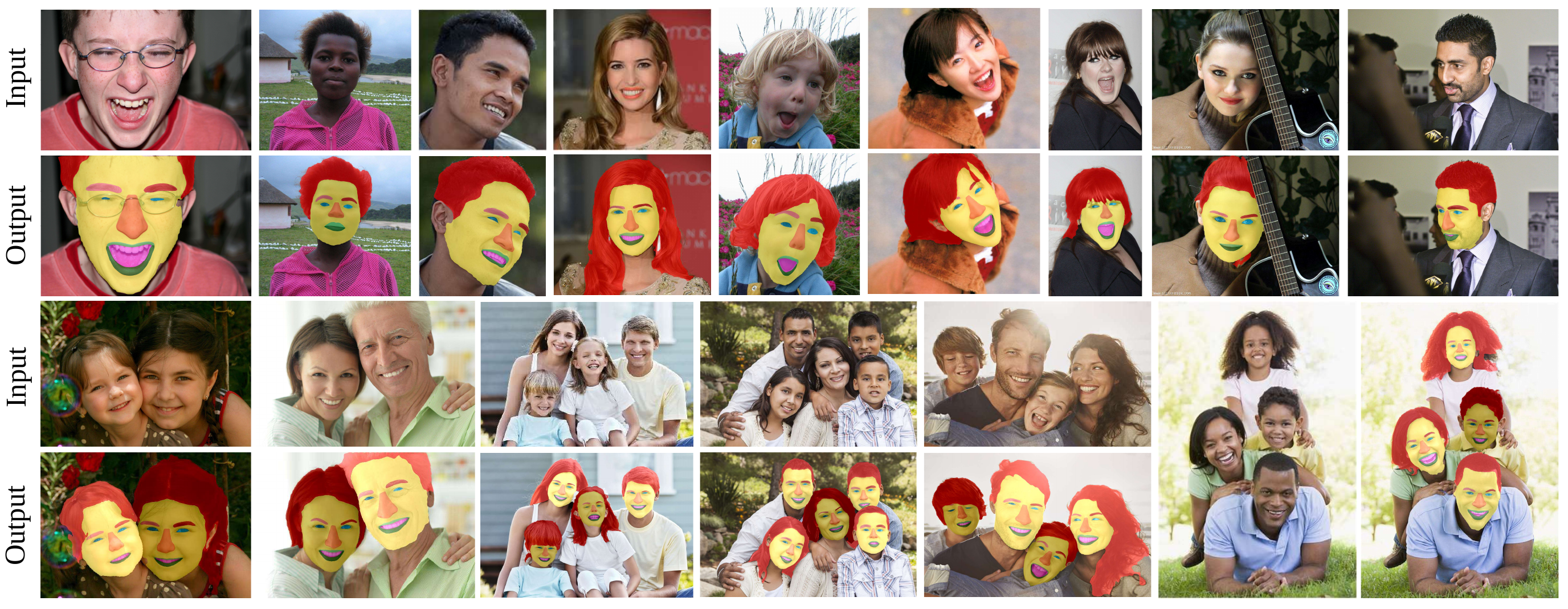}
\caption{Qualitative results on challenging images. Adjacent individuals are distinguished by different brightnesses. (Best viewed in color)}
\label{fig:challenging}
\vspace{-0.7em}
\end{figure*}

\noindent\textbf{Importance of Tanh-warping:}
Models that apply cropping $\mathbb{C}$ or Tanh-wrapping $\mathbb{W}$ using face rectangles consistently outperform the baseline $\texttt{rescale}$+\emph{FCN} model (Table \ref{table:compare_with_baseline_helen_star} Row 1) on all labels. The key factor is that both $\mathbb{C}$ and $\mathbb{W}$ operators can allow the network to focus on a smaller but more important region by amplifying its resolution. It evidences the effectiveness of the \emph{central vision}.
It is not guaranteed, however, that the central vision always covers all true regions. For example, using $\mathbb{C}$ on an input image may crop off hairs distant from the face, as shown by Figure \ref{fig:sigmoid_warp_invert}. By combining the peripheral vision with the central vision together, Tanh-warping $\mathbb{W}$ brings significant improvements in F-measure on hairs over $\mathbb{C}$ (88.8 vs. 85.0).

Cropping a larger region seems to be a reasonable alternative to Tanh-warpping since it increases the hair coverage of the network input. However, it conversely decreases the resolution of the face, causing the segmentation performance to drop. We add a result called $\mathbb{C}2$ with 2 times larger cropping area in Table \ref{table:compare_with_baseline_helen_star}. $\mathbb{C}2$ is even worse than $\mathbb{C}$ for either outer components (\emph{hair} 83.8 vs. 84.5) or inner components (\emph{overall} 90.9 vs. 92.9). Therefore, a larger crop region is not a good solution to the limited region problem caused by cropping.

\noindent\textbf{Importance of Hybrid Structure:}
The proposed hybrid structure combines a FCN with a Mask R-CNN-like structure. The  Mask R-CNN part adopts the central vision for inner facial components by applying RoI-align in the feature level. Table \ref{table:compare_with_baseline_helen_star} illustrates the advantages of \emph{Hybrid} structures over pure \emph{FCN} structures on the overall scores for inner facial components (92.0 vs. 89.4, 91.9 vs. 89.3). Padding the regressed bounding boxes for RoI-align further improves the overall scores (92.9 vs. 92.0, 93.1 vs. 91.9). Meanwhile, we observe that our \emph{Hybrid} structure does not affect the hair scores much compared with \emph{FCN} (88.7 vs. 88.8).

\noindent\textbf{Importance of Separated Segmentation Modules:}
Our hybrid network consists of multiple segmentation modules for different inner facial components. Different from the mask heads of Mask R-CNN, our segmentation modules do not share weights. The importance of separated weights is verified by the results from Table \ref{table:compare_with_baseline_helen_star}, which shows that sharing weights across all component segmentation modules for inner facial components reduces the overall accuracy (92.7 vs. 93.1). 
The eye/brow symmetric weight sharing is comparable to separated weights (\emph{eyes} 89.7 vs. 89.7; \emph{brows} 85.8 vs. 85.9; \emph{overall} 93.0 vs. 93.1).
We also train an official Mask R-CNN on HELEN* where facial components are detected through a region proposal network. As explained in Section \ref{ssec:component_prediction}, it does not perform well for face parsing, which is verified by the scores (\emph{hair} 78.5, \emph{overall} 84.9).

\subsection{More Results}

\noindent\textbf{Varying hair lengths:} Due to the nonlinear rescaling ability of Tanh-warping, as shown in Figure \ref{fig:hairs}, our method is suitable for segmenting hairs with various lengths. 

\noindent\textbf{In-the-wild and multi-face conditions:} Figure~\ref{fig:challenging} shows visual results on challenging images from HELEN datasets and Internet. Although our model is trained on HELEN* dataset, it shows the ability to handle large pose, expression variations, occlusion and multiple closely-snuggled faces in the wild condition. 
More concretely, for the multi-face scenario,
we first detect 5-point landmarks for each face appearing in the image, then apply the RoI Tanh-warping and hybrid network for each face independently, and de-warp its softmax activation maps onto the original image through bilinear mapping. Finally, on each pixel, its instance is determined by selecting the maximal foreground activation score among all faces.
These results in Figure \ref{fig:challenging} show that our method is capable of distinguishing different face instances.

\noindent\textbf{Efficiency:} The proposed network is efficient. It runs at $40$ms per face on Nvidia Titan Xp GPU.

\section{Conclusion}
We propose a novel hybrid network combined with RoI Tanh-warping for face parsing with hairs. We use RoI Tanh-warping to align the face in the middle while preserving the peripheral context for parsing hairs. Our hybrid network applies a Mask R-CNN-fashion branch for inner components (eyes, brows, nose, mouth), while applying a FCN-fashion branch for outer components (face, hair). Ablation studies show the effectiveness of RoI Tanh-warping and our hybrid structure. The superior performances on public datasets HELEN/LFW-PL and in-the-wild images show the ability of our method to handle the problem of face parsing with hairs under various environments.

\section*{Acknowledgments}
This work is partially supported by NSFC~(No. 61402387) and Guiding Project of Fujian Province, China~(No. 2018H0037).

{\small
\bibliographystyle{ieee}
\bibliography{face}
}

\end{document}